# Robust Noise Filtering in Image Sequences

Soumaya Hichri
Signal, Image processing and Patterns Recognition
Laboratory ENIT-Tunisia
Campus Universitaire, B.P.37, 1002 Tunis, Belvédère, Tunisie

Faouzi Benzarti
Signal, Image processing and Patterns Recognition
Laboratory ENIT-Tunisia
Campus Universitaire, B.P.37, 1002 Tunis, Belvédère, Tunisie

Hamid Amiri
Signal, Image processing and Patterns Recognition
Laboratory ENIT-Tunisia
Campus Universitaire, B.P.37, 1002 Tunis, Belvédère, Tunisie

## ABSTRACT
Image sequences filtering have recently become a very important technical problem especially with the advent of new technology in multimedia and video systems applications. Often image sequences are corrupted by some amount of noise introduced by the image sensor and therefore inherently present in the imaging process. The main problem in the image sequences is how to deal with spatio-temporal and non stationary signals. In this paper, we propose a robust method for noise removal of image sequence based on coupled spatial and temporal anisotropic diffusion. The idea is to achieve an adaptive smoothing in both spatial and temporal directions, by solving a nonlinear diffusion equation. This allows removing noise while preserving all spatial and temporal discontinuities.

## 1. INTRODUCTION
Any acquired images either: photographic, astronomical, medical or others are subject to noise degradation which reduces the quality of the image. The noise is introduced due to thermal fluctuations in sensors, quantization effects, and properties of communication channels. It affects the perceptual quality of the image decreasing not only the appreciation of the image but also the performance of the task for which the image has been intended. The challenge in the image filtering is to design methods, which can selectively smooth a degraded image or image sequence without altering edges and losing significant features. In the case of still image, the filtering acts in a space of two dimensions (2D), whereas in a sequence images (or video image), the temporal component is added (2D+t) and should be taken into account. To effectively adapt the local structures of the image and to preserve discontinuities, it should be recommended to the use of a non linear filter. Recently, a new approach based on Partial Differential Equations (PDEs) [1][2], has emerged as a more powerful and a successfully approach for studying a variety of problems including image restoration, image segmentation, and image denoising. The PDE methods have remarkable advantages in both theory and computation. They allow handling and processing visually important geometric features such as gradients, tangents, curvatures and modeling visually meaningful dynamic process such as linear and nonlinear diffusion. In our approach, we use the concept of PDEs to formulate an anisotropic diffusion model. The key idea is to perform an adaptive smoothing which allowing isotropic smoothing in low gradients (e.g. homogeneous area), while avoiding diffusion in high gradients (e.g. discontinuities), thus preserving the natural edge of the image. For image sequences, we have to consider both spatial and temporal gradients. We distinguish three classes of filters: static filters, adaptive filters and motion compensated filters. Static filters are obtained by extending the 2D image filter support to the 3D time-space support [3]. We call them static because they do not take into account the dynamic character of image sequences. Adaptive filters are designed specially to deal with image sequences and take into account the possibility of a motion [4][5]. Motion compensated filters estimate explicitly the motion of the sequence by a motion estimation algorithm [6]. It is assumed that, in order to deal with the dynamic character of sequences and to obtain high quality results, motion estimation is necessary.

This paper is organized as follow: In section 2, we review the mathematical foundation of the PDEs. In section 3, we present the proposed method and discuss the estimation of the model's parameters. In the last section, we exhibit the experimental results.

## 2. PDE-BASED IMAGE DENOISING
In this section, we will review the mathematical formalism of the PDEs, and the concept of isotropic and anisotropic diffusion applied to the still grayscale image.

### 2.1 Isotropic diffusion
The isotropic diffusion finds its origin in the well known heat equation according to the following equation [7]:

$$\begin{cases} \frac{\partial g(x,y,t)}{\partial t} = \Delta g(x,y,t) = div(\nabla g(x,y,t)) \\ g(x,y,0) = f(x,y) \quad (Initial\ condition) \end{cases}$$
(1)

Where t: time of diffusion; $\nabla g$: is the gradient of the grayscale image g,
The parabolic linear PDE (1) can be interpreted as a diffusion process of pixel (x,y)'s brightness around the neighboring pixels (x±∂x, y±∂y), during a time t (t∈[0,T]). It is shown that this approach is not efficient, because the diffusion operates in all directions leading to edge degradation.

### 2.2 Anisotropic diffusion
To overcome the problem of the isotropic diffusion, the anisotropic diffusion has been proposed [1][8], in which smoothing is only performed in low gradients areas (homogeneous areas), the idea is to introduce a decreasing positive function $c(|\nabla g|)$ (i.e. diffusion function) controlling the force of the diffusion making it possible to distinct noise/edges. This leads to the following expression:

$$\begin{cases} \frac{\partial g}{\partial t} = div(c(|\nabla g|\nabla g) \\ g(x,y,0) = f(x,y) \quad [I.C] \\ |\nabla g| = \sqrt{(g_x^2 + g_y^2)} \end{cases} \quad (2)$$

Where $g_x$, $g_y$: first derivative components of g while posing $c(\nabla g) = \frac{\Phi'(|\nabla g|)}{|\nabla g|}$, equation (2) can be written as [9]:

$$\begin{cases} \frac{\partial g}{\partial t} = \Phi''(|\nabla g|)g_{\eta\eta} + \frac{\Phi'(|\nabla g|)}{|\nabla g|}g_{\xi\xi} \\ g(x,y,0) = f(x,y) \quad [I.C] \\ \frac{\partial g}{\partial \eta}\Big|_{\partial\Omega} = 0 \quad [B.C] \end{cases} \quad (3)$$

With $g_{\eta\eta}$ and $g_{\xi\xi}$ are the second order directional derivatives respectively along the gradient direction $\eta = \frac{\nabla f}{|\nabla f|}$ and along its orthogonal direction $\xi = \eta^{\perp}$.

The formulation (3) allows defining conditions on $\Phi(.)$ function, in order to obtain a process stable and converging:

i) $\Phi''(|\nabla g|) \geq 0 \ et \ \Phi'(|\nabla g|) \geq 0$
ii) $\lim_{|\nabla g|\to 0}\frac{\Phi'(|\nabla g|)}{|\nabla g|} = \lim_{|\nabla g|\to 0}\Phi''(|\nabla g|) = \Phi''(0) > 0$
iii) $\lim_{|\nabla g|\to\infty}\frac{\Phi'(|\nabla g|)}{|\nabla g|} = 0, \lim_{|\nabla g|\to\infty}\Phi''(|\nabla g|) \ et \ \lim_{|\nabla g|\to\infty}\frac{\Phi''(|\nabla g|)}{\frac{\Phi'(|\nabla g|)}{|\nabla g|}} = 0$

(4)

Condition (*i*) avoids inverse diffusion along $\eta$ and $\xi$. Condition (*ii*) allows isotropic diffusion for low gradient, with no preferred diffusion directions since $\eta$ and $\xi$ do not represent significant orientations. Condition (*iii*) allows anisotropic diffusion to preserve discontinuities for the high gradient. Among functions satisfying conditions -*i*) -*ii*) –*iii*), the Perona function [4], while posing $u=|\nabla f|$:

$$\Phi(u,k) = k^2/2(1 - e^{-(u/k)^2}) \quad (5)$$

This function introduces a parameter *k* or diffusity which acts as a threshold which determines whether to preserve edges or not. Areas in which the gradient magnitude is lower than *k* will be blurred more strongly than areas with a higher gradient magnitude. This tends to smooth uniform regions, while preserving the edges between different regions.

## 2.3 Anisotropic diffusion in the case of image sequence

The most obvious solution is to perform anisotropic diffusion frame-per-frame in the image sequences. But in this case redundancies between successive frames wouldn't be exploited, and those provide very useful information for noise removal. The idea is to extend the 2D equation (2) to 3D by defining a 3D gradient to detect discontinuities. Thus, the PDE can be reformulated as follows:

$$\begin{cases} \frac{\partial g(x,y,z,t)}{\partial t} = div\begin{pmatrix} c_s(|\nabla g|_s)\frac{\partial g}{\partial x} \\ c_s(|\nabla g|_s)\frac{\partial g}{\partial y} \\ c_s(|\nabla g|_s)\frac{\partial g}{\partial z} \end{pmatrix} \\ g(x,y,z,0) = f(x,y,z) \quad [I.C] \\ |\nabla g|_s = \sqrt{g_x^2 + g_y^2 + g_z^2} \end{cases} \quad (6)$$

Where z: denotes the temporal component, $c_s(.)$: the spatial diffusion function, Perona-Malik propose for example the function expressed by equation (7), and $|\nabla g|_s$ the gradient norm 3D.

$$c(|\nabla g|) = \exp\left(-\frac{|\nabla g|^2}{k^2}\right) \quad (7)$$

This method seems effective to noise removal but has some drawbacks: the gradient norm 3D doesn't give any information on the variation either it spatial or temporal. With consequent, the correlations between frames of the same sequence (i.e. temporal information) will not be well exploited. We can show them on the figure 1.
From these results, we note that the main problem connected with noise filtering of image sequences is how to deal with the spatio-temporal gradient.

## 3. PROPOSED METHOD

The proposed method is derived from the anisotropic diffusion (3D) process (6), except that the spatial and temporal discontinuities have their own detector. The idea is to perform simultaneously two operations of non linear filters: one in the space domain, and the other in the time domain. In figure 2, we present the adopted model. From noisy sequence $g(i,j,n)$ (i.e. discrete form of $g(x,y,z)$ as input, both spatial and temporal gradient norm are estimated and then incorporated in the anisotropic PDE to denoise the degraded image sequence. In the following sub-section, we will give some details about the PDE model and the temporal gradient estimation. The parameters of the model are adjusted manually.

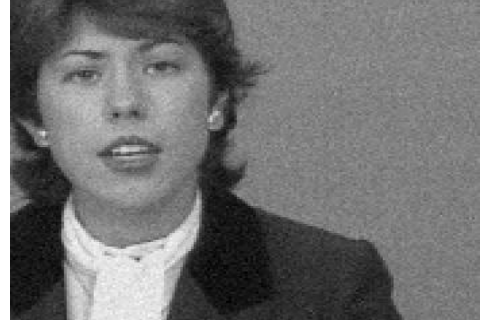

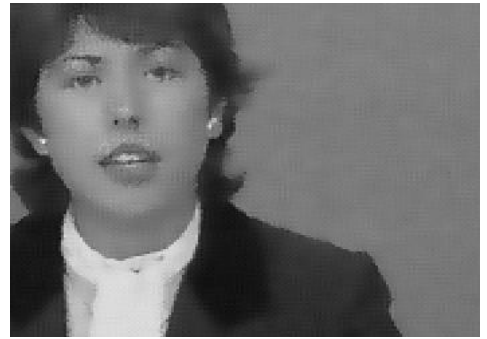

**Fig 1: Anisotropic diffusion (3D) of noisy sequence "claire" (frame n°5), -a- noisy sequence, -b- filtered sequence**

## 3.1 Coupled spatial and temporal anisotropic diffusion

As mentioned above, our method is based on coupling spatial and temporal anisotropic diffusion, leading to the following PDE [10]:

$$\begin{cases} \frac{\partial g}{\partial t} = \underbrace{(g-f)}_{terme\ 1} + \lambda\ div \underbrace{\begin{pmatrix} c_s(|\nabla g|_s)\frac{\partial g}{\partial x} \\ c_s(|\nabla g|_s)\frac{\partial g}{\partial y} \\ c_t(|\nabla g|_t)\frac{\partial g}{\partial z} \end{pmatrix}}_{terme\ 2} \\ g(x,y,z,0) = f(x,y,z) \qquad [I.C] \end{cases}$$
(8)

The first term in equation (8) ensures fidelity of the data; the second term corresponds to the constraint of regularization. The parameter λ ensures the compromise between the two terms. $c_s(.)$ the spatial diffusion function, $c_t(.)$ the temporal diffusion function, $|\nabla g|_s$ representing spatial discontinuities, and $|\nabla g|_t$ representing temporal discontinuities.

The discrete equation of (8) is given by:

$$g^{n+1}(i,j) = g^n(i,j) + \Delta t\left[(g^n(i,j) - f^n(i,j)) + \lambda\left[\sum_{p\in\eta s} c_s(|\nabla g|_s)\nabla g^n_{s,p} + \sum_{p'\in\eta t} c_t(|\nabla g|_t)\nabla g^n_{s,p'}\right]\right]$$
9)

Where $\Delta t$: the parameter ensuring numerical stability of the PDE generally between $0 \leq \Delta T \leq 0.24$,
$\nabla g_{s,p} = g_p - g_s$ and $\nabla g_{s,p'} = g_{p'} - g_s$ with $s,p \in \eta s$ and $s,p' \in \eta t$. With $\eta s$ and $\eta t$ respectively the spatial and temporal neighborhood of the pixel $s$.

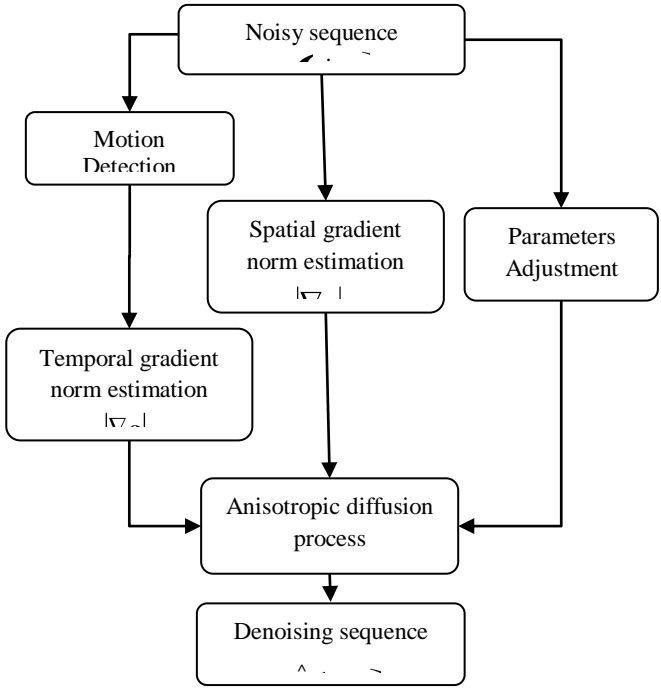

Fig 2: Proposed model

## 3.2 Temporal gradient estimation

The easiest solution would be to compute a 1D gradient calculated on coordinate $z$ for each pixel of this sequence. Unfortunately, the high noise sensibility of this method makes it unusable. We propose using a robust method based on the Σ−Δ modulation which is used in electronic field for Analog to Digital converter (ADC), for encoding high resolution signals into lower resolution signals using pulse density modulation [11][12].

In our case, we interpret background estimation as the simulation of a digital conversion of a time-varying analog signal. In our case, we interpret background estimation as the simulation of a digital conversion of a time-varying analog signal. We can show on the figure 3.

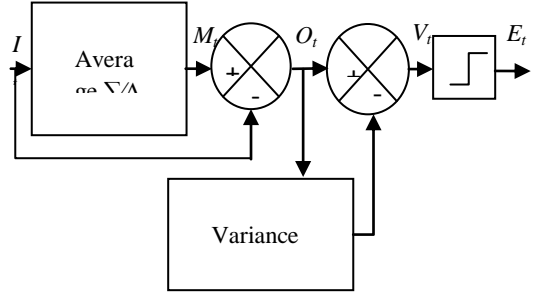

Fig 3: modulator Σ / Δ for a sequence of images

Where: $I_t$ is the input sequence, $M_t$ is the estimated background value, $O_t$ is the difference image, $V_t$ is image of variance calculated for each pixel and $E_t$ is the image of binary labels (movement/background).

So, at every image, the estimate is simply incremented by one if it is smaller than the sample or decremented by one if it is greater than the sample.

The main steps of the Σ−Δ algorithm for temporal gradient estimation are as follow:

- Step (1): computes the Σ−Δ mean.

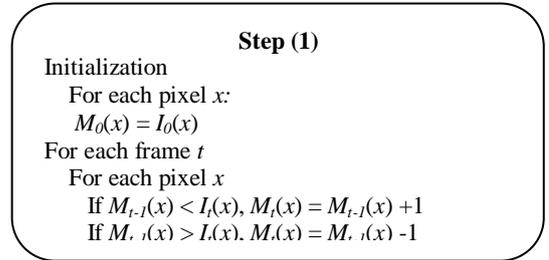

- Step (2): Computes of the difference between the image and the Σ−Δ mean (motion likelihood measure).

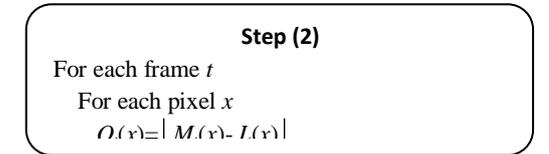

- Step (3): Computes of the Σ−Δ variance defined as the Σ−Δ mean of N times the non-zero differences.

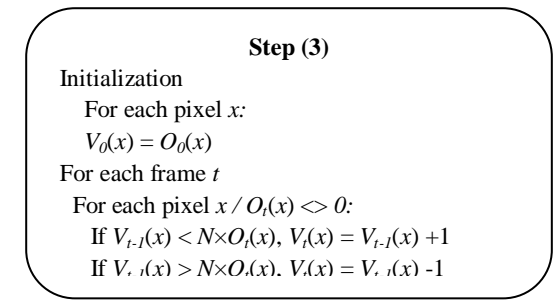

- Step (4): Computes of the motion label by comparison between the difference and the variance.

**Step (4)**
For each frame *t*
  For each pixel *x:*
    If $O_t(x) < V_t(x)$
      Then $E_t(x) = 0$  // *static pixel*
      Else $E_t(x) = 1$  // *movment pixel*

## 4. EXPERIMENTAL RESULTS

In this section, we present the experimental results. In the first experimental, we test the performance of the algorithm with method of diffusion 3D and with median method.

The initialized parameters are fixed to: *N*=450 (the number of iterations) and $\Delta t = .24$. The regularization parameters values are fixed to: k=0.21, λ=0.53. The denoising sequence in figure 4, shows a significantly improvement: edges and discontinuities have been recovered and preserved with a good suppression of noise.

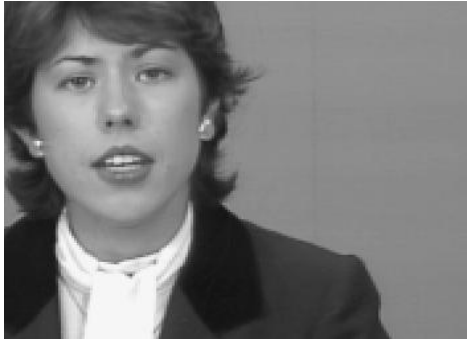

-a-

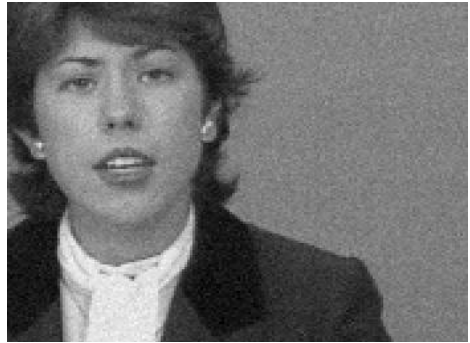

-b-

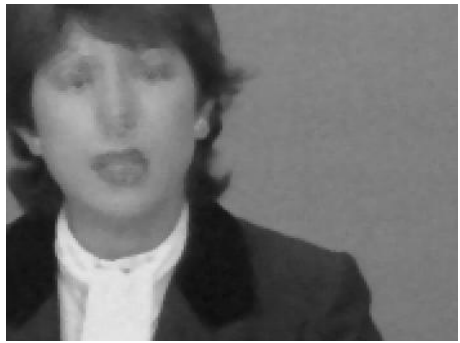

-c-

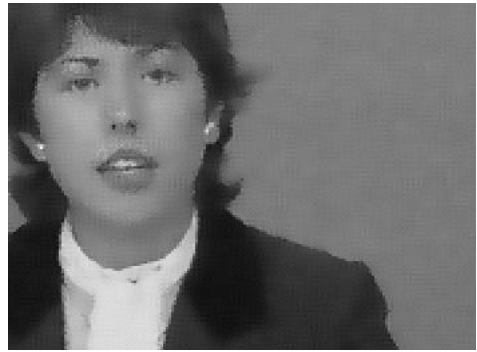

-d-

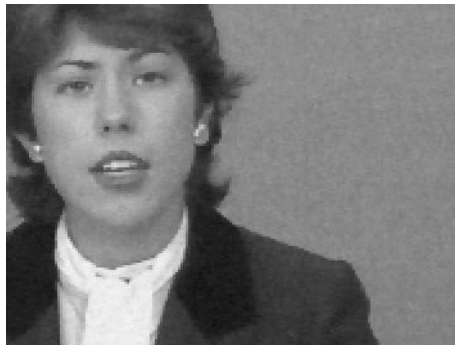

-e-

**Fig 4: Noise removal of "Claire" sequence (frame5)** -a-*Original test sequence, -b- noised sequence, -c- sequence denoised with Median3D method,-d- diffusion method 3D, -e- Proposed method*

We note that the proposed method has the best PSNR value as shown in Table 1. Whereas, PSNR an objective measure to evaluate the image quality, defined by:

$$PSNR(n) = 10 * \log(D^2/MSE(n)) \quad (10)$$

Where *MSE*: the mean square error between the original sequence and the estimated one. D: the maximum pixel value in the image.

$$MSE(n) = (1/M*N)\left(\sum_{i=1}^{M}\sum_{j=1}^{N}\left(f(i,j,n) - \hat{f}(i,j,n)\right)\right) \quad (11)$$

Where:
$f(i,j,n)$: n[th] frame of original noisy sequence.
$\hat{f}(i,j,n)$: n[th] frame of denoising estimated sequence.
*M*N*: Size of frame.

**Table 1 . Values of EQM and PSNR**

| Methods | EQM(5) | PSNR(5) |
|---|---|---|
| Proposed method | 0 ,0007 | 31,5079 |
| Diffusion 3D | 0,0030 | 25,1797 |
| Method Median | 0,0065 | 21,8709 |

Figure 5 shows the variation of PSNR according to image of the sequence for the different methods.

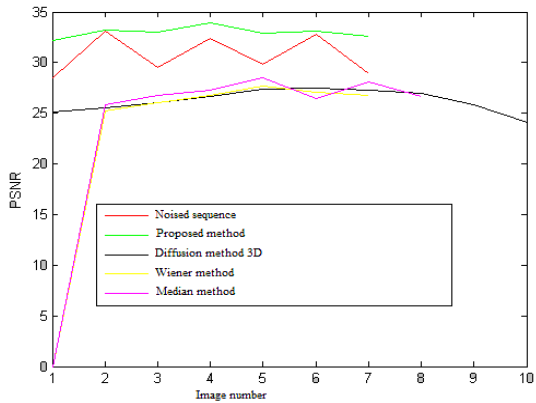

**Fig 5: Denoising sequence by different methods**

From these results, we can see that the proposed model provides better image quality with significantly reduced noise and discontinuity preserving. Also compared to other methods, our model has the largest values of PSNR. Clearly the results as viewed by the other methods are less satisfactory compared to our method of coupled anisotropic diffusion filtering.

## 5. CONCLUSION

In this paper, we have proposed a new approach of denoising image sequence using a coupled spatial and temporal diffusion via the Partial Differential Equations (PDE). The key idea behind anisotropic approach is to incorporate an adaptive smoothness. That is, the diffusion is encouraged in the homogeneous region and discourage across boundaries. This new approach should contribute to reduce noise, and preserving spatio-temporal edges frames. To estimate temporal gradient, we incorporate the Sigma-Delta technique which is very efficient to detect temporal discontinuities.The results are very promising, and open new prospects for the treatment by EDP for image sequences. Future work will include automatic diffusion/detection parameters estimation that would lead to an unsupervised method.